\pdfoutput=1

\documentclass[11pt]{article}

\usepackage[final]{acl}

\usepackage{times}
\usepackage{latexsym}

\usepackage[T1]{fontenc}

\usepackage[utf8]{inputenc}

\usepackage{microtype}

\usepackage{inconsolata}

\usepackage{graphicx}

\usepackage{amsthm,amsmath,amssymb}
\usepackage{bbding}
\usepackage{float}
\usepackage{bm}
\usepackage{booktabs}
\usepackage{graphicx}
\usepackage{multirow}
\usepackage{diagbox}
\usepackage{enumitem}
\usepackage{setspace}
\usepackage{algorithm}
\usepackage{algorithmic}

\title{Investigating the (De)Composition Capabilities of Large Language Models in Natural-to-Formal Language Conversion}

\author{Ziyao Xu, Houfeng Wang \\
  National Key Laboratory for Multimedia Information Processing, \\
  School of Computer Science, Peking University \\
  \texttt{\{xzyxzy,wanghf\}@pku.edu.cn}}

\begin{document}
\maketitle
\begin{abstract}
To achieve generalized and robust natural-to-formal language conversion (N2F), large language models (LLMs) need to have strong capabilities of decomposition and composition in N2F when faced with an unfamiliar formal language and be able to cope with compositional gaps and counter-intuitive symbolic names. To investigate whether LLMs have this set of basic capabilities in N2F, we propose the DEDC framework. This framework semi-automatically performs sample and task construction, allowing decoupled evaluation of the set of decomposition and composition capabilities of LLMs in N2F. Based on this framework, we evaluate and analyze the most advanced LLMs, and the main findings include that: (1) the LLMs are deficient in both decomposition and composition; (2) the LLMs show a wide coverage of error types that can be attributed to deficiencies in natural language understanding and the learning and use of symbolic systems; (3) compositional gaps and counter-intuitive symbolic names both affect the decomposition and composition of the LLMs. Our work provides a new perspective for investigating the basic capabilities of decomposition and composition of LLMs in N2F. The detailed analysis of deficiencies and attributions can help subsequent improvements of LLMs. The dataset and code are available at \url{https://github.com/xzy-xzy/DEDC}.%

\end{abstract}

\section{Introduction}
A formal language \cite{salomaa1987formal} is a symbolic system formed by a set of symbolic primitives that can be combined into expressions to convey specific meanings according to certain rules. Common formal languages include formal logic \cite{prior1963formal}, structured query language \cite{DBLP:conf/sigmod/ChamberlinB74}, formal syntax \cite{sag1999syntactic}, etc. Formal languages are often used to more accurately convey the meaning embedded in natural language texts, resulting in many natural-to-formal language conversion (N2F) tasks in the field of natural language processing, such as syntactic-semantic parsing \cite{DBLP:journals/corr/abs-2006-11056}, structured query language generation \cite{DBLP:journals/vldb/KatsogiannisMeimarakisK23}, logical expression generation for symbolic reasoning \cite{DBLP:conf/acl/Xu0P0LH24}, etc.

Due to the wide coverage of the training corpus, large language models (LLMs) have developed, to some extent, the capability of N2F on a variety of common formal languages \cite{DBLP:conf/acl/LiuCSZNH0L24}. However, it is an under-explored question whether LLMs have developed basic capabilities to cope with arbitrary new formal languages for N2F. When confronted with an unfamiliar formal language, the capabilities of \textit{decomposition} and \textit{composition} is necessary to accomplish N2F: after seeing some samples containing expressions in formal language and their corresponding meanings in natural language, LLMs need to be able to decompose the meanings of symbolic primitives from the samples (\textit{decomposition}) and combine the symbolic primitives into new expressions with specific meanings (\textit{composition}). In addition, to achieve generalized and robust N2F, LLMs need to have the capability to cope with the following situations during decomposition and composition: (1) there are compositional gaps between the expressions to be combined and the expressions seen, and (2) there are names of symbolic primitives that would not normally correspond to the actual meanings of the symbolic primitives (counter-intuitive). %

\begin{figure*}[t]
    \centering
    \includegraphics[width=1.0\textwidth]{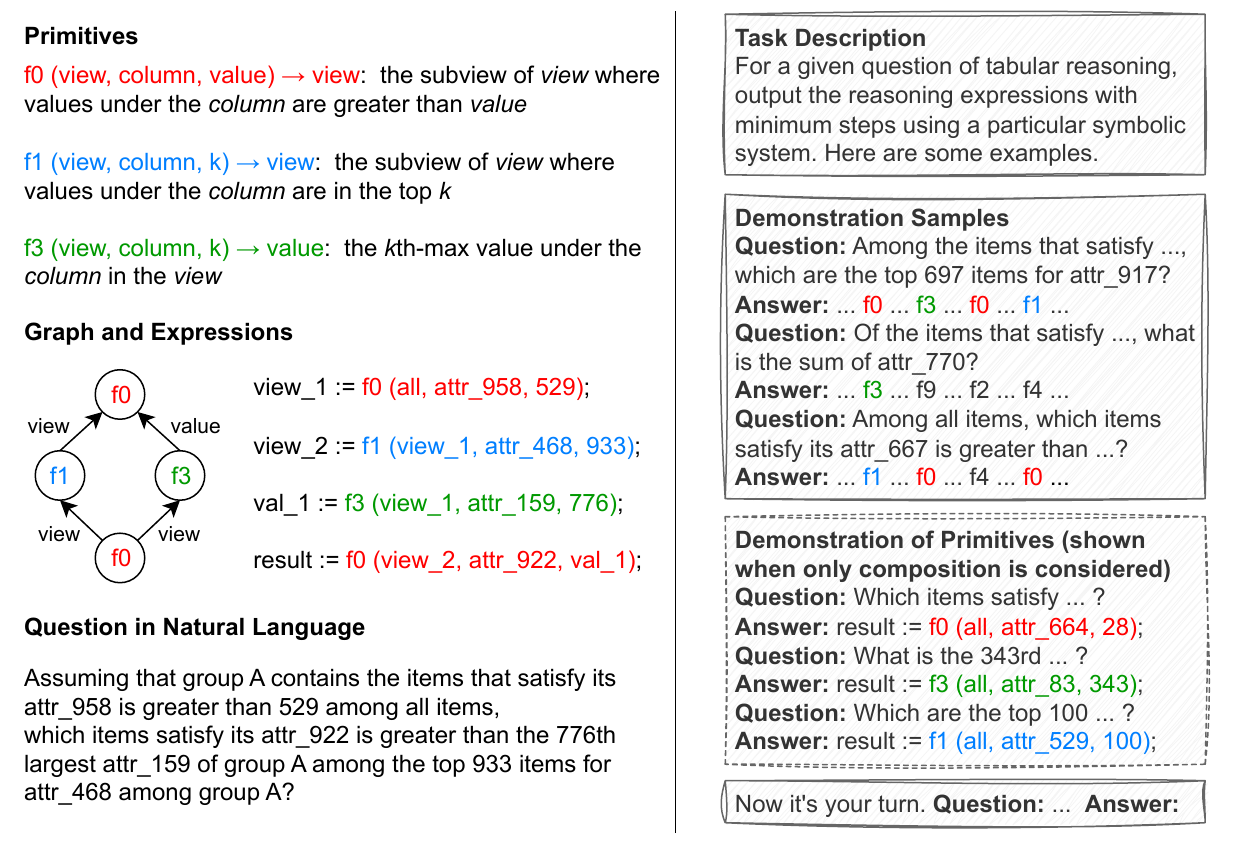}
    \caption{An illustration of the sample and task construction of the DEDC framework. An example of the sample construction is shown on the left. On the right is an illustration of the task construction when the sample on the left is used as the test sample.}
    \label{fig:framework}
\end{figure*}

Investigating whether LLMs have this set of capabilities is important from both practical and cognitive perspectives \cite{DBLP:journals/corr/abs-2210-03050}. From a practical perspective, this set of capabilities is necessary for LLMs to accomplish N2F under in-context learning \cite{dong2024surveyincontextlearning}, especially when confronted with uncommon formal languages and complex situations. From a cognitive perspective, the intelligence reflected in this set of capabilities is one of the necessary conditions to confirm that LLMs are truly intelligent \cite{10.1145/1283920.1283930} in N2F. Evaluation and analysis are needed to determine whether LLMs have this set of capabilities and to find deficiencies for subsequent improvement. However, there is currently no framework to support such evaluation and analysis.

To solve this problem, we propose the DEDC framework. This framework constructs samples and tasks in a semi-automatic manner, allowing \textbf{D}ecoupled \textbf{E}valuation of LLMs' \textbf{D}ecomposition and \textbf{C}omposition capabilities in N2F. In addition, this framework allows for easy setting adaptations to evaluate the capabilities of LLMs to cope with compositional gaps and counter-intuitive symbolic names in N2F.

Based on the DEDC framework, we conduct a comprehensive evaluation and analysis of the above-mentioned decomposition and composition capabilities of existing LLMs. We also conduct a detailed analysis of the specific error behaviors of the LLMs in N2F to provide more insights into the attributions of errors. %

\section{DEDC Framework}

\subsection{Sample Construction}

Figure~\ref{fig:framework} shows an example of sample construction on the left.

\textbf{Formal language.} We consider a formal language with 10 primitives in the context of tabular reasoning \cite{tabfact, chen-etal-2020-logic2text} as the case for investigation, where the primitives are functions and the expressions are assignment statements using a single function. By combining multiple expressions, the output of one function can be used as one of the input arguments to other functions, thus expressing complex tabular reasoning processes. See Appendix~\ref{app:prim} for details of the primitives.

\begin{figure}
    \centering
    \includegraphics[width=\columnwidth]{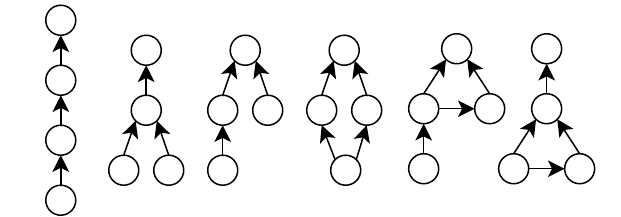}
    \caption{Six types of base graphs we identify for the sample construction of the STD framework.}
    \label{fig:base_graph}
\end{figure}

\textbf{Graph and expressions.} The compositional structure of expressions can be represented as a directed acyclic graph \cite{mcd}, where the nodes are functions and the directed edges indicate that the output of one function is used as an input parameter to another function. We first identify 6 types of base graphs that contain 4 nodes (shown in Figure~\ref{fig:base_graph}) and then enumerate all the schemes that identify a function for each node, yielding a total of 323 valid schemes. For each valid scheme, we randomly generate structure-independent parameters to obtain multi-step expressions corresponding to the scheme. See Appendix~\ref{app:scheme} for details of scheme enumeration.

\textbf{Question in natural language.} All valid schemes correspond to 18 types of graphs with output types on the edges. For each type of graph, we manually designed question generation templates to generate the corresponding natural language questions from the expressions. See Appendix~\ref{app:question} for details of question generation. See Appendix~\ref{app:gram} for a discussion of grammatical divergence in question generation.

With the above construction process, we obtain 323 samples of (graph, expressions, question).

\subsection{Task Construction}
\par We use each sample as a test sample for task construction. The right side of Figure~\ref{fig:framework} shows an example of task construction when the sample on the left is used as a test sample.

\par For each test sample, the task of LLMs is to transform the input question into the corresponding expressions after seeing three demonstration samples. The demonstration samples are randomly selected from samples other than the test samples and satisfy the following conditions:
\begin{itemize}[itemsep=2pt,topsep=2pt,parsep=2pt]

\item[$\bullet$] Each demonstration sample contains at least one primitive in the test sample.

\item[$\bullet$] The set of all primitives in the three demonstration samples covers the set of primitives in the test sample.
\end{itemize}

Demonstration samples are enough for LLMs to learn desired primitives and basic rules of composition. To accomplish this task, LLMs need both decomposition and composition capabilities: they need to decompose the meaning and format of the primitives from the demonstration samples (decomposition), and then combine the primitives according to the question to obtain expressions with the corresponding meaning (composition). We denote the performance of a LLM on the task as $P_{dc}$.

\subsection{Metric}

We use accuracy ($\%$) as the performance metric. A directed acyclic graph may have multiple topological orders and thus correspond to multiple correct orders of expressions. When determining whether an answer is correct, we traverse the expressions in order and perform variable substitution to obtain a final single expression and determine whether it agrees with the standard answer. This approach avoids the influence of multiple topological orders and the arbitrariness of intermediate variable names on the determination. In addition, if the expression contains primitives with swappable arguments, we check the final expression before and after the argument swapping, and consider it correct if one of them agrees with the standard answer. See Appendix~\ref{app:answer} for details of answer checking.

\subsection{Decoupling}

To decouple the effects of decomposition and composition capabilities of the LLM on this task, we also measure the performance of the LLM on the task when it can see the demonstration of primitives (shown in Figure~\ref{fig:framework}), which provides a corresponding simplest sample for each primitive in the test sample. In this case, the LLM can directly obtain the meaning and format of the desired primitives without decomposition capabilities, and only needs composition capabilities to accomplish the task. We denote the performance of the LLM in this case as $P_c$. After measuring $P_{dc}$ and $P_c$ separately, we can estimate the effects of decomposition and composition capabilities of the LLM on task performance based on $P_{dc}$ and $P_c$:
\begin{itemize}[itemsep=2pt,topsep=2pt,parsep=2pt]

\item[$\bullet$] We use $D_{c} = 100 \ - \ P_{c}$ to estimate the effect of composition capability. $D_c$ indicates the error rate of the LLM when only composition is needed. A larger $D_c$ indicates a larger defect in the composition capacity of the LLM.

\item[$\bullet$] We use $D_{d} = P_{c} - P_{dc}$ to estimate the effect of decomposition capability. $D_d$ indicates the additional error rate of the LLM due to the need for decomposition. A larger $D_d$ indicates a larger defect in the decomposition capacity of the LLM.
\end{itemize}

In this way, the effects of decomposition and composition capabilities of the LLM are decoupled and comparable.

\section{Base Evaluation}  %

In this section, based on the DEDC framework, we perform a base evaluation (i.e., no additional settings) of the LLMs and analyze the results.

\subsection{Models}
We select a series of the most advanced LLMs for evaluation, including the closed-source LLMs GPT-4o-20240806 \cite{openai2024gpt4technicalreport}, Claude-3.5-sonnet-20240620 \cite{claude}, and the open-source LLMs DeepSeek-2.5 \cite{deepseekai2024deepseekllmscalingopensource}, Mistral-large-2407 \cite{jiang2024mixtralexperts}, and Llama-3.1-405B \cite{dubey2024llama3herdmodels}. To get the deterministic output, we set the temperature parameter to $0$.

\subsection{Results}

Table~\ref{tab:base-res} shows the results of the base evaluation. From the results we find that:

(1) The LLMs are relatively good at composition but still have deficiencies. All LLMs except DeepSeek-2.5 have $D_c<10$, i.e., the error rate is less than $10\%$ when only composition is required, reflecting a relatively strong composition capability. However, no LLMs can achieve perfect performance with zero error rate, and most LLMs show a wide coverage of error types (we will discuss this in detail in~\ref{subsec:error_analysis}), indicating that there are still deficiencies in composition.

(2) The LLMs have significant deficiencies in decomposition, which are more severe than in composition. All LLMs except Claude-3.5 have $D_d>10$, i.e., the additional error rate due to need decomposition exceeds $10\%$. Even for Claude-3.5, the additional error rate reaches $7.74\%$. This indicates that the LLMs have severe deficiencies in decomposition. All LLMs have $D_d>D_c$, suggesting that the deficiencies in decomposition are more severe than in composition.

(3) Closed-source LLMs in general have greater decomposition and composition capabilities than open-source LLMs. The closed-source model outperforms the open-source model in all comparisons except that Mistral-large has a slightly lower $D_c$ than GPT-4o. The closed-source model Claude-3.5 shows the strongest decomposition and composition capabilities.

(4) Except for Mistral-large, the $D_d$ ranking of the LLMs is consistent with the $D_c$ ranking, showing a correlation between decomposition and composition capabilities to some extent. As an exception, Mistral-large shows relatively strong composition capability but weak decomposition capability.

\begin{table}[]
\resizebox{\columnwidth}{!}
{
\begin{tabular}{l|cccc}
\hline
              & $P_{dc}$ & $P_c$ & $D_c$ & $D_d$ \\ \hline
GPT-4o        & 81.42    & 94.74 & 5.26  & 13.31 \\
Claude-3.5    & \textbf{91.02}   & \textbf{98.76} & \textbf{1.24}  & \textbf{7.74}  \\
DeepSeek-2.5  & 68.73    & 86.69 & 13.31 & 17.96  \\
Mistral-large & 76.78    & 95.98 & 4.02  & 19.20 \\
Llama-3.1     & 74.92    & 90.71 & 9.29  & 15.79 \\ \hline
\end{tabular}
}
\caption{\label{tab:base-res} Results of the base evaluation of the LLMs.}
\end{table}

\begin{table*}[t]

\centering

\begin{tabular}{p{0.1\textwidth}|p{0.4\textwidth}|p{0.4\textwidth}}
\hline
\textbf{Type} & \textbf{Sample 1} & \textbf{Sample 2} \\
\hline

Primitive confusion
&
{
\par \textbf{Ans:} view\_1 := \textcolor{blue}{f0} (all, attr\_641, 684);
\par \textbf{Err:} view\_1 := \textcolor{red}{f2} (all, attr\_641, 684);
\par \textbf{Note:} f0 and f2 have different input formats but the same output type
}
&
{
\par \textbf{Ans:} value\_2 := \textcolor{blue}{f3} (view\_2, attr\_611, 25);
\par \textbf{Err:} row\_1 := \textcolor{red}{f6} (view\_2, attr\_611, 25);
\par \textbf{Note:} f3 and f6 have different output types but the same input format
}
\\ \hline

Primitive fiction
&
{
\par \textbf{Ans:} view\_1 := \textcolor{blue}{f0 (all, attr\_814, 380)}; value\_1 := \textcolor{blue}{f3 (view\_1, attr\_175, 342)};
\par \textbf{Err:} value\_1 := \textcolor{red}{f3 (all, attr\_814, 380, attr\_175, 342)};
\par \textbf{Note:} Compositional meaning
}
&
{
\par \textbf{Ans:} view\_2 := \textcolor{blue}{f2 (view\_1, attr\_346, attr\_486)};
\par \textbf{Err:} view\_2 := \textcolor{red}{f2 (all, attr\_346, attr\_486)}; view\_3 := \textcolor{red}{f6 (view\_1, view\_2)};
\par \textbf{Note:} Independent new meaning
}
\\ \hline

Variable misuse
& 
{
\par \textbf{Ans:} \textcolor{blue}{value\_1}:= f5 (all); view\_1 := f0 (all, attr\_7, \textcolor{blue}{value\_1}); 
\par \textbf{Err:} \textcolor{red}{view\_1} := f5 (all); view\_2 := f0 (all, attr\_7, \textcolor{red}{value\_1});
\par \textbf{Note:} Non-existent variable name makes the second expression invalid
} 
&
{
\par \textbf{Ans:} \textcolor{blue}{view\_3} := f1 (view\_2, attr\_267, 65); result := f2 (\textcolor{blue}{view\_3}, attr\_941, attr\_825);
\par \textbf{Err:} \textcolor{red}{view\_3} := f1 (view\_2, attr\_267, 65); result := f2 (\textcolor{red}{view\_2}, attr\_941, attr\_825);
\par \textbf{Note:} Incorrect variable name makes the first expression useless
}
\\ \hline

Redundan-cy
&
{
\par \textbf{Ans:} value\_2 := \textcolor{blue}{f3 (view\_2, attr\_81, 362)};
\par \textbf{Err:} row\_1 := \textcolor{red}{f6 (view\_2, attr\_81, 362)}; value\_2 := \textcolor{red}{f7 (row\_1, attr\_81)};
\par \textbf{Note:} Equivalent but redundant
}
&
{
\par \textbf{Ans:} ... view\_1 := f2 (\textcolor{blue}{all}, attr\_892, col\_1);
\par \textbf{Err:} \textcolor{red}{view\_1 := f2 (all, attr\_892, attr\_87);} ... view\_2 := f2 (\textcolor{red}{view\_1}, attr\_892, col\_1);
\par \textbf{Note:} Incorrect redundant steps
}
\\ \hline

Omission
&
{
\par \textbf{Ans:} \textcolor{blue}{view\_1 := f2 (all, attr\_675, attr\_210);} value\_1 := f4 (\textcolor{blue}{view\_1}, attr\_690);
\par \textbf{Err:} value\_1 := f4 (\textcolor{red}{all}, attr\_690);
\par \textbf{Note:} Omission of scope
}
&
{
\par \textbf{Ans:} \textcolor{blue}{col\_1 := f9 (attr\_221, value\_1);} view\_1 := f2 (all, attr\_27, \textcolor{blue}{col\_1});
\par \textbf{Err:} view\_1 := f2 (all, attr\_27, \textcolor{red}{value\_1});
\par \textbf{Note:} Omission of operation
}
\\ \hline

Incorrect meaning
&
{
\par \textbf{Ans:} view\_1 := ...; view\_2 := ...; value\_1 := f4 (\textcolor{blue}{view\_1}, attr\_716); view\_3 := f0 (\textcolor{blue}{view\_2}, attr\_319, value\_1);
\par \textbf{Err:} view\_1 := ...; view\_2 := ...; value\_1 := f4 (\textcolor{red}{view\_2}, attr\_716); view\_3 := f0 (\textcolor{red}{all}, attr\_319, value\_1);
\par \textbf{Note:} Incorrect antecedents
}
&
{
\par \textbf{Ans:} view\_1 := \textcolor{blue}{f1} (all, attr\_511, 512); value\_1 := \textcolor{blue}{f3} (view\_1, attr\_651, 345); view\_2 := \textcolor{blue}{f0} (all, attr\_896, value\_1);
\par \textbf{Err:} value\_1 := \textcolor{red}{f3} (all, attr\_651, 345); view\_1 := \textcolor{red}{f0} (all, attr\_896, value\_1); view\_2 := \textcolor{red}{f1} (view\_1, attr\_511, 512);
\par \textbf{Note:} Confusing ordering
}
\\ \hline
\end{tabular}

\caption{\label{tab:error} Six types of errors that occur in LLMs in the base evaluation and corresponding examples. \textbf{Ans} refers to the standard answer, and \textbf{Err} refers to the LLM output with errors (only part of the expressions are shown). We mark the main differences between the standard answer and the LLM output in \textcolor{blue}{blue} and \textcolor{red}{red}.}
\end{table*}

\subsection{Error Analysis}
\label{subsec:error_analysis}

We check the specific error behaviors of the LLMs and summarize the following six error types (see Table~\ref{tab:error} for examples):

\textbf{Primitive confusion}. The LLMs use the correct input parameters but the incorrect primitive name in an expression.

\textbf{Primitive fiction}. The LLMs fictionalize a non-existent primitive and apply the name of an existing primitive to it. The fictionalized primitive combines the meanings of several existing primitives or has an independent new meaning.

\textbf{Variable misuse}. The LLMs use non-existent or incorrect intermediate variable names, resulting in invalid or useless expressions.

\textbf{Redundancy}. The LLMs use redundant expressions which result in an output that is incorrect or does not have minimum steps.

\textbf{Omission}. The LLMs ignore a certain part of the natural language question, resulting in the missing of necessary expressions.

\textbf{Incorrect meaning}. The meaning of the expressions output by the LLMs is not consistent with the meaning of the natural language question. From the perspective of graphs, there is an inconsistency between the graph corresponding to the output and the graph of the sample, such as incorrect antecedents for some nodes or confusing ordering of nodes.

Of the six error types, \textbf{Incorrect meaning} and \textbf{Omission} can be completely attributed to conversion errors caused by the LLMs' misunderstanding of the meaning of natural language questions. In error types \textbf{Primitive confusion} and \textbf{Primitive fiction}, the LLMs do not incorrectly understand the meaning that needs to be expressed, but make errors in the use of formal language, which can be completely attributed to deficiencies in the capability to learn and use the symbolic system. Error types \textbf{Variable misuse} and \textbf{Redundancy} contain instances where both attributions mentioned above make sense. The error types suggest that both the capability of natural language understanding and the capability of learning and using symbolic systems should be considered in N2F.

\begin{table}[t]

\centering

\begin{tabular}{l|cccccc}
\hline
Error in $P_c$ & Pc & Pf & Vm & R & O & Im \\ \hline
\multicolumn{1}{l|}{GPT-4o} & 8 & 3 & 1 & 1 & 1 & 3 \\
\multicolumn{1}{l|}{Claude-3.5} & 2 & - & 2 & - & - & - \\
\multicolumn{1}{l|}{DeepSeek-2.5} & 30 & 2 & - & 4 & 1 & 6 \\
\multicolumn{1}{l|}{Mistral-large} & 7 & 2 & 1 & 1 & - & 2 \\
\multicolumn{1}{l|}{Llama-3.1} & 14 & 1 & 7 & 1 & - & 7 \\ \hline
\multicolumn{1}{l}{} \\ \hline
Error in $P_{dc}$ & Pc & Pf & Vm & R & O & Im \\ \hline
\multicolumn{1}{l|}{GPT-4o} & 41 & 5 & 5 & 3 & 2 & 4 \\
\multicolumn{1}{l|}{Claude-3.5} & 24 & 1 & 3 & - & 1 & - \\
\multicolumn{1}{l|}{DeepSeek-2.5} & 71 & 9 & - & 6 & 3 & 12 \\
\multicolumn{1}{l|}{Mistral-large} & 66 & 4 & 1 & 1 & 1 & 4 \\
\multicolumn{1}{l|}{Llama-3.1} & 52 & 6 & 7 & 7 & - & 9 \\ \hline

\end{tabular}

\caption{\label{tab:error_m} The number of occurrences of each error type for the LLMs when only composition is required (above) and when both decomposition and composition are required (below). \  - \ indicates that the corresponding error type does not occur.} %
\end{table}

For each of the LLMs, we count the number of occurrences of each error type, and the results are shown in Table~\ref{tab:error_m}. From the results we find that:

(1) The LLMs show a wide coverage of error types. When only composition is required, the LLMs already show a wide coverage of error types even at relatively low error rates: except for Claude-3.5, which shows only 2 error types, all the other LLMs show 5 $\sim$ 6 error types. When decomposition is required, the LLMs show more error types or more instances under some error types. The wide coverage of error types suggests that the LLMs have many pending deficiencies in both composition and decomposition in N2F.

(2) The most common error type for the LLMs is \textbf{Primitive confusion}. Even without the need to decompose the meaning and format of the primitives, \textbf{Primitive confusion} is the most frequent error type for each of the LLMs. When decomposition is required, the number of error instances of \textbf{Primitive confusion} also grows the most. This suggests that the deficiencies of the LLMs in learning and using the symbol system in N2F are severe.

(3) Different LLMs have different characteristics in the distribution of error types. Except for \textbf{Primitive confusion}, common error types vary across the LLMs. Some of the LLMs do not show certain error types, e.g., Claude-3.5 does not show error types \textbf{Incorrect meaning} and \textbf{Redundancy}. The characteristics in the distribution of error types can help subsequent analysis and targeted improvements of the LLMs in N2F.

\section{Evaluations with Additional Settings}

In the DEDC framework, additional settings can be added to evaluate the capability of LLMs to cope with a variety of situations and estimate the effects of the additional settings on the decomposition and composition of the LLMs. Assuming that the performance of a LLM on the task with and without demonstration of primitives under setting $s$ is $P^{s}_{c}$ and $P^{s}_{dc}$ respectively, we estimate the effect of the setting by comparing the performance with that of the base evaluation:

\begin{itemize}[itemsep=2pt,topsep=2pt,parsep=2pt]

\item[$\bullet$] We use $\triangle^{s}_{c} = P^{s}_{c} - P_{c}$ to estimate the effect of setting $s$ on composition of the LLM. $\triangle^{s}_{c}$ indicates the change in the performance of the LLM caused by setting $s$ when only composition is needed.

\item[$\bullet$] We use $\triangle^{s}_{d} = (P^{s}_{dc} - P_{dc}) - \triangle^{s}_{c}$ to estimate the effect of setting $s$ on decomposition of the LLM. $\triangle^{s}_{d}$ indicates the additional performance change of the LLM caused by setting $s$ due to the need for decomposition.
\end{itemize}

Under this definition, $\triangle^{s}_{c}>0$ means that the setting makes composition of the LLM easier, $\triangle^{s}_{c}<0$ means that the setting makes composition of the LLM more difficult, and $|\triangle^{s}_{c}|$ indicates the magnitude of the effect. The meaning of $\triangle^{s}_{d}$ in decomposition of the LLM is similar.

In this section, we consider settings related to compositional gaps and counter-intuitive symbolic names.

\subsection{Compositional Gap}
\begin{figure}
    \centering
    \includegraphics[width=\columnwidth]{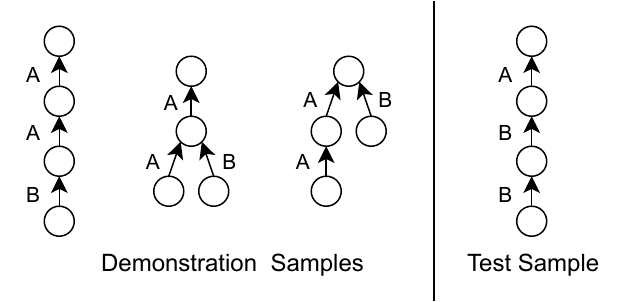}
    \caption{An illustration of the compositional gap between the test sample and its demonstration samples. A and B indicate the different output types on the edges.}
    \label{fig:gap}
\end{figure}

\subsubsection{Settings}

Compositional generalization is a common type of generalization in language-related tasks is considered to be one of the key linguistic manifestations of human intelligence \cite{kim-linzen-2020-cogs, DBLP:journals/jair/HupkesDMB20, kim-linzen-2020-cogs, xu-wang-2024-spor}. After learning some samples, humans can cope with test samples formed by recombining elements in the learned samples, even if they have not seen the compositional form of the test samples during learning (i.e., there are compositional gaps between the test samples and the learned samples). In N2F, the relatively limited primitive space of formal languages makes more situations that require compositional generalization. To achieve generalized N2F, LLMs need to have the capability to handle compositional gaps.

In the DEDC framework, we regard the primitives as elements and consider the capability of LLMs to cope with the compositional gaps between the demonstration samples and the test samples. The compositional form of primitives is determined by the graphs with output types on the edges in the sample, so we define that a test sample has a compositional gap with its demonstration samples if and only if any graph with output types on the edges in the demonstration samples is different from that in the test sample (shown in Figure~\ref{fig:gap}).

In the base evaluation, 70\% of the test samples have compositional gaps with their demonstration samples. We consider the following two settings for compositional gaps:

\begin{itemize}[itemsep=2pt,topsep=2pt,parsep=2pt]

\item[$\bullet$] 0\% gap. Any test sample does not have a compositional gap with its demonstration samples.

\item[$\bullet$] 100\% gap. Each test sample has a compositional gap with its demonstration samples.
\end{itemize}

We achieve the settings by changing the demonstration samples for samples that do not satisfy the requirements in the base evaluation.

\begin{table}[]
\resizebox{\columnwidth}{!}
{
\begin{tabular}{lcccc}
\hline
\multicolumn{1}{l|}{0\% gap}       & $P^{s}_{dc}$         & $P^{s}_{c}$          & $\triangle^{s}_{c}$  & $\triangle^{s}_{d}$  \\ \hline
\multicolumn{1}{l|}{GPT-4o}        & 86.69                & 96.59                & +1.86                & +3.41                \\
\multicolumn{1}{l|}{Claude-3.5}    & \textbf{94.43}                & \textbf{99.38}                & +0.62                & +2.79                \\
\multicolumn{1}{l|}{DeepSeek-2.5}  & 76.16                & 93.19                & +6.50                & +0.93                \\
\multicolumn{1}{l|}{Mistral-large} & 83.28                & 98.45                & +2.48                & +4.02                \\
\multicolumn{1}{l|}{Llama-3.1}     & 83.59                & 94.74                & +4.02                & +4.64                \\ \hline
                                   & \multicolumn{1}{l}{} & \multicolumn{1}{l}{} & \multicolumn{1}{l}{} & \multicolumn{1}{l}{} \\ \hline
\multicolumn{1}{l|}{100\% gap}     & $P^{s}_{dc}$         & $P^{s}_{c}$          & $\triangle^{s}_{c}$  & $\triangle^{s}_{d}$  \\ \hline
\multicolumn{1}{l|}{GPT-4o}        & 79.26                & 93.19                & -1.55                & -0.62                \\
\multicolumn{1}{l|}{Claude-3.5}    & \textbf{89.78}                & \textbf{98.14}                & -0.62                & -0.62                \\
\multicolumn{1}{l|}{DeepSeek-2.5}  & 63.47                & 84.52                & -2.17                & -3.10                \\
\multicolumn{1}{l|}{Mistral-large} & 76.47                & 95.98                & 0                 & -0.31                \\
\multicolumn{1}{l|}{Llama-3.1}     & 71.83                & 87.62                & -3.10                & 0                 \\ \hline
\end{tabular}
}
\caption{\label{tab:gap-res} Results of the evaluation under the setting of 0\% compositional gap (above) and 100\% compositional gap (below).}
\end{table}

\subsubsection{Results and Analysis}
Table~\ref{tab:gap-res} shows the results of the evaluation under the two settings for compositional gaps. All the LLMs show $\triangle^{s}_{c}>0$, $\triangle^{s}_{d}>0$ under the setting of 0\% gap and $\triangle^{s}_{c}\leq 0$, $\triangle^{s}_{d}\leq 0$ under the setting of 100\% gap. This indicates that the composition and decomposition performance of the LLMs decreases as the proportion of samples with compositional gaps increases, i.e., compositional gap makes the composition and decomposition of the LLMs more difficult.

Additional errors due to compositional gaps encompass all six types described in~\ref{subsec:error_analysis}, suggesting that the effect of compositional gaps on the LLMs is relevant to both natural language understanding and the learning and use of symbolic systems. The effect of compositional gaps on composition can be attributed to the difficulties of the LLMs in dealing with compositional forms that have not been seen before; the effect on decomposition suggests that the decomposition of the LLMs is not independent of the test samples and can be disturbed by compositional gaps between the test samples and the demonstration samples.

\subsection{Counter-intuitive Symbolic Name}

\subsubsection{Settings}

The nature of the symbolic system depends on the meanings of the symbolic primitives and does not change as the names of the symbolic primitives change. Achieving robust N2F requires that LLMs have an invariance: for a set of symbolic primitives with established meanings, there is no significant difference in their performance under different symbolic names. This invariance essentially requires that LLMs be able to think based on symbolic meanings independently of symbolic names, and simply use the names in their expressions.

\begin{figure}
    \centering
    \includegraphics[width=\columnwidth]{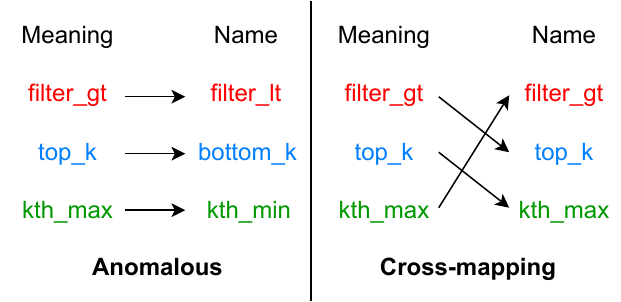}
    \caption{An illustration of the two types of settings for counter-intuitive symbolic names. The arrow indicates that a primitive with the meaning on the left end uses the name on the right end.}
    \label{fig:counter}
\end{figure}

In the DEDC framework, the primitives can be named arbitrarily. To investigate whether LLMs have such invariance in N2F, we evaluate the performance of LLMs under the setting of counter-intuitive symbolic names, which refers to names of symbolic primitives that would not normally correspond to the actual meanings of the primitives. If a LLM can think independently of symbolic names and use symbolic names correctly in its expression, then counter-intuitive symbolic names should not affect its performance in N2F. We consider the following two types of settings for counter-intuitive symbolic names (see Figure~\ref{fig:counter} for examples):

\begin{itemize}[itemsep=2pt,topsep=2pt,parsep=2pt]

\item[$\bullet$] \textbf{Anomalous}. For each primitive, use the name of another primitive that has the same form as the primitive but a completely different meaning. For example, use \textsl{bottom\_k} as the name of a primitive with the meaning \textsl{top\_k}.

\item[$\bullet$] \textbf{Cross-mapping}. For each primitive, use the name that normally corresponds to the meaning of another primitive in the framework. For example, use \textsl{kth\_max} as the name of a primitive with the meaning \textsl{top\_k}.
\end{itemize}

We achieve the settings by simply replacing the names of the primitives in the base evaluation.

\subsubsection{Results and Analysis}
Table~\ref{tab:counter-res} shows the results of the evaluation under the two types of settings for counter-intuitive symbolic names. All the LLMs show $\triangle^{s}_{c}<0$, $\triangle^{s}_{d}<0$ under the two types of settings, suggesting that both types of counter-intuitive symbolic names, \textbf{Anomalous} and \textbf{Cross-mapping}, make the composition and decomposition of the LLMs more difficult. The effect of counter-intuitive symbolic names is more severe compared to compositional gaps, especially on decomposition, as all the LLMs show $|\triangle^{s}_{d}|>20$ which indicates that the settings increase the additional error rate due to the need for decomposition by more than 20\%. On composition, there are significant differences between the effects of different types of settings on different LLMs, e.g., Llama-3.1 is almost unaffected by the \textbf{Anomalous} setting, but Mistral-large is severely affected; all LLMs except Mistral-large are more affected by the \textbf{Cross-mapping} setting than the \textbf{Anomalous} setting.

The additional error due to counter-intuitive symbolic names centers on the misuse of symbolic names: the LLMs use intuitive symbolic names (e.g., the Meaning column of Figure~\ref{fig:counter}) instead of the symbolic names indicated in the demonstration samples (e.g., the Name column of Figure~\ref{fig:counter}). This concentrated type of error suggests that counter-intuitive symbolic names primarily influence the LLMs' use of symbolic names in expressions rather than their thought processes. We hypothesize that under the paradigm of generating the next token based on maximum probability, the LLMs cannot well handle the conflict between intuitive symbolic names and contextual guidance, leading to the incorrect use of symbolic names in expressions. When only composition is required, the demonstration of primitives mitigates this conflict but does not eliminate it; when decomposition is additionally required, the absence of demonstration of primitives makes the conflict more severe. Intuitive symbolic names may appear in the demonstration samples under the \textbf{Cross-mapping} setting but not in the \textbf{Anomalous} setting, so the \textbf{Cross-mapping} setting usually leads to more severe conflicts; however, LLMs that are inherently familiar with intuitive symbolic names may suffer more severe conflicts under the \textbf{Anomalous} setting. 

\begin{table}[]
\resizebox{\columnwidth}{!}
{
\begin{tabular}{lcccc}
\hline
\multicolumn{1}{l|}{\textbf{Anomalous}}     & $P^{s}_{dc}$         & $P^{s}_{c}$          & $\triangle^{s}_{c}$  & $\triangle^{s}_{d}$  \\ \hline
\multicolumn{1}{l|}{GPT-4o}                 & 52.94                & 87.00                & -7.74                & -20.74               \\
\multicolumn{1}{l|}{Claude-3.5}             & \textbf{57.28}                & \textbf{95.36}                & -3.41                & -30.34               \\
\multicolumn{1}{l|}{DeepSeek-2.5}           & 41.80                & 80.50                & -6.19                & -20.74               \\
\multicolumn{1}{l|}{Mistral-large}          & 13.93                & 68.11                & -27.86               & -34.98               \\
\multicolumn{1}{l|}{Llama-3.1}              & 45.82                & 90.09                & -0.62                & -28.48               \\ \hline
                                            & \multicolumn{1}{l}{} & \multicolumn{1}{l}{} & \multicolumn{1}{l}{} & \multicolumn{1}{l}{} \\ \hline
\multicolumn{1}{l|}{\textbf{Cross}} & $P^{s}_{dc}$         & $P^{s}_{c}$          & $\triangle^{s}_{c}$  & $\triangle^{s}_{d}$  \\ \hline
\multicolumn{1}{l|}{GPT-4o}                 & 44.27                & 84.52                & -10.22               & -26.93               \\
\multicolumn{1}{l|}{Claude-3.5}             & \textbf{53.56}                & \textbf{92.57}                & -6.19                & -31.27               \\
\multicolumn{1}{l|}{DeepSeek-2.5}           & 33.13                & 72.45                & -14.24               & -21.36               \\
\multicolumn{1}{l|}{Mistral-large}          & 26.63                & 73.99                & -21.98               & -28.17               \\
\multicolumn{1}{l|}{Llama-3.1}              & 35.29                & 75.54                & -15.17               & -24.46               \\ \hline
\end{tabular}
}
\caption{\label{tab:counter-res} Results of the evaluation under the setting of \textbf{Anomalous} (above) and \textbf{Cross-mapping} (below) for counter-intuitive symbolic names.}
\end{table}

\section{Related Work}

\noindent \textbf{(De)Composition}. Research on composition emerges before the era of LLMs \cite{pmlr-v80-lake18a, DBLP:journals/jair/HupkesDMB20}, focusing on compositional gaps between training and test sets \cite{kim-linzen-2020-cogs, mcd}. In the era of LLMs, some work investigates composition under in-context learning by sampling from pre-partitioned datasets with compositional gaps \cite{levy-etal-2023-diverse, an-etal-2023-context}. One notion of decomposition in the research is the decomposition of the steps of task execution \cite{8777144}. Another notion of decomposition concerns the decomposition of specific contents, and related work involves the decomposition of large-scale data and complex problems for reasoning \cite{10.1145/3539618.3591708} and planning \cite{wu2024mldt}, and grammar-related decomposition tasks such as morphological analysis \cite{moisio2024llmsmorphologicalanalysescomplex}. In this work, we try to investigate the decomposition and composition of primitives in N2F and avoid pre-partitioning in evaluation.

\noindent \textbf{N2F.} Related work under the N2F topic covers a wide range of formal language types, such as syntax for linguistic analysis \cite{biaffine, shi-etal-2024-structured}, machine-executable languages \cite{shaw-etal-2021-compositional, ma2024sciagenttoolaugmentedlanguagemodels}, etc. The language-specific N2F capabilities of LLMs can often be obtained from a targeted training corpus \cite{jiang2024surveylargelanguagemodels, shi2024surveyemployinglargelanguage}. Our work provides a new perspective that investigates whether LLMs possess the basic capabilities of decomposition and composition in N2F.

\section{Conclusion}

In this work, we propose the DEDC framework. This framework performs sample and task construction semi-automatically, allowing decoupled evaluation of the decomposition and composition capabilities of LLMs in N2F. In addition, this framework allows for the evaluation of the capability of LLMs to cope with compositional gaps and counter-intuitive symbolic names. Based on this framework, we evaluate and analyze the most advanced LLMs. From the results we find that: (1) the LLMs are deficient in both decomposition and composition, and the deficiencies are more severe in decomposition; (2) the LLMs show a wide coverage of error types, indicating that the deficiencies are relevant to both natural language understanding and the learning and use of symbolic systems; (3) the LLMs do not cope well with compositional gaps and counter-intuitive symbolic names, and both decomposition and composition are affected. Our work provides a new perspective for investigating the basic capabilities of decomposition and composition of LLMs in N2F. Our analysis of error types and attributions can help subsequent improvements of LLMs, and the DEDC framework can be used for evaluation and analysis of the improvements.

\section*{Limitations}

\par In this work, we concentrate on the evaluation and detailed analysis of LLMs in a single case of decomposition and composition in N2F. This leads to two limitations. First, the cases covered by our work are not comprehensive enough, mainly in terms of the types of formal languages we investigate. Second, our work does not provide specific methods for improving the decomposition and composition capabilities of LLMs. Nevertheless, we believe that our investigative work provides a new perspective on the decomposition and composition capabilities of LLMs in N2F and some meaningful findings. The DEDC framework can be generalized to some extent to a wider range of cases (see Appendix~\ref{app:gen} for a discussion), and the evaluation results and analysis can help subsequent research on improvements. Based on this work, we will try to cover more cases and investigate improvement methods in our future work.

Another limitation of this work is the lack of human-related investigations from a cognitive perspective. See Appendix~\ref{app:human} for a discussion related to human capabilities.

\section*{Ethics Statement}

\par We comply with the license to use LLMs for scientific research purposes only. The datasets we construct will also be open source for scientific research purposes. The datasets we use and construct do not contain any information that names or uniquely identifies individual people or offensive content.
\par The AI assistant we use in our work is Copilot (for simple code completion).

\section*{Acknowledgements}

This work was supported by National Science and Technology Major Project (No. 2022ZD0116308) and the Academic Research Projects of Beijing Union University (No. ZK10202405). The corresponding author is Houfeng Wang.

\bibliography{custom}

\clearpage
\appendix

\begin{table*}[h]

\centering

\begin{tabular}{p{0.37\textwidth}|p{0.57\textwidth}}
\hline
\textbf{Name \& Format} & \textbf{Meaning} \\ \hline
f0 (view, column, value) → view & the subview of \textit{view} where values under the \textit{column} are greater than \textit{value} (\textsl{filter\_gt}) \\ \hline
f1 (view, column, k) → view & the subview of \textit{view} where values under the \textit{column} are in the top \textit{k} (\textsl{top\_k}) \\ \hline
f2 (view, column1, column2) → view & the subview of \textit{view} where values under the \textit{column1} are greater than values under \textit{column2} (\textsl{filter\_gt\_c}) \\ \hline
f3 (view, column, k) → value & the \textit{k}th-max value under the \textit{column} in the \textit{view} (\textsl{kth\_max}) \\ \hline
f4 (view, column) → value & the sum of values under the \textit{column} in the \textit{view} (\textsl{sum}) \\ \hline
f5 (view) → value & the number of rows in the \textit{view} (\textsl{count}) \\ \hline
f6 (view, column, k) → row & the row with the \textit{k}th-max value under the \textit{column} in the \textit{view} (\textsl{kth\_argmax}) \\ \hline
f7 (row, column) → value & the value of the \textit{column} in the \textit{row} (\textsl{hop}) \\ \hline
f8 (value1, value2) → value & \textit{value1} + \textit{value2} (\textsl{add}) \\ \hline
f9 (column, value) → column & The \textit{column} in which all values increase by \textit{value} (\textsl{add\_c}) \\ \hline
\end{tabular}

\caption{\label{tab:prim} The primitives used in the DEDC framework.}
\end{table*}

\section{Framework Details}

\subsection{Primitives}
\label{app:prim}
Table~\ref{tab:prim} shows the 10 primitives used in the DEDC framework. We avoid using multiple primitives that play the same structural role in the graphs, as this leads to an unbalanced distribution of structures. In the base evaluation, we use primitive names without any meaning information attached to them to avoid that LLMs have ever seen the same or similar primitive name-meaning mappings.

\subsection{Scheme Enumeration}
\label{app:scheme}
A scheme is obtained by enumerating a primitive for each node on a base graph without restriction. A valid scheme must satisfy that each predecessor node $Y$ of a node $X$ can match an unique input parameter of $X$ with the same type as the output type of $Y$. In addition to this, we exclude schemes that satisfy any of the following conditions:

(1) There are consecutive math operation nodes or consecutive filter nodes. This can result in schemes that potentially correspond to multiple different execution paths.

(2) The predecessor of a count node is a single top\_k node. This can lead to meaningless operations.

(3) The final node is a math operation node. This will result in the corresponding question not having a practical meaning.

(4) There is a pair of nodes with the same predecessor and successor, and the node with the smaller ordinal number has the primitive with the larger label. This can lead to schemes that are isomorphic to other schemes.

\subsection{Question Generation}
\label{app:question}

\begin{table*}[h]

\centering

\begin{tabular}{p{0.03\textwidth}|p{0.22\textwidth}|p{0.65\textwidth}}
\hline
 & \textbf{Attr / Func} & \textbf{String} \\ \hline
\multirow{4}{*}{f0} & link\_scope & "among" \\ \cline{2-3}
   & leaf\_scope( ) & "the items that satisfy its \{self.h\} is greater than \{self.val\} among all items" \\ \cline{2-3}
   & scope(val=self.val) & "the items that satisfy its \{self.h\} is greater than \{val\}" \\ \cline{2-3}
   & question(val=self.val) & "which items satisfy its \{self.h\} is greater than \{val\}" \\ \hline

\multirow{4}{*}{f1} & link\_scope & "among" \\ \cline{2-3}
   & leaf\_scope( ) & "the largest \{self.k\} items for \{self.h\} among all items" \\ \cline{2-3}
   & scope( ) & "the largest \{self.k\} items for \{self.h\}" \\ \cline{2-3}
   & question( ) & "which are the largest \{self.k\} items for \{self.h\}" \\ \hline

\multirow{4}{*}{f2} & link\_scope & "among" \\ \cline{2-3}
   & leaf\_scope( ) & "the items that satisfy its \{self.h\} is greater than its \{self.col\} among all items" \\ \cline{2-3}
   & scope(val=self.col) & "the items that satisfy its \{self.h\} is greater than \{val\}" \\ \cline{2-3}
   & question(val=self.col) & "which items satisfy its \{self.h\} is greater than \{val\}" \\ \hline

\multirow{4}{*}{f3} & link\_scope & "of" \\ \cline{2-3}
   & leaf\_value( ) & "the \{self.order\} largest \{self.h\} of all items" \\ \cline{2-3}
   & value( ) & "the \{self.order\} largest \{self.h\}" \\ \cline{2-3}
   & question( ) & "what is the \{self.order\} largest \{self.h\}" \\ \hline

\multirow{4}{*}{f4} & link\_scope & "of" \\ \cline{2-3}
   & leaf\_value( ) & "the sum of \{self.h\} of all items" \\ \cline{2-3}
   & value( ) & "the sum of \{self.h\}" \\ \cline{2-3}
   & question( ) & "what is the sum of \{self.h\}" \\ \hline

\multirow{4}{*}{f5} & link\_scope & "of" \\ \cline{2-3}
   & leaf\_value( ) & "the number of all items" \\ \cline{2-3}
   & value( ) & "the number" \\ \cline{2-3}
   & question( ) & "what is the number" \\ \hline

\multirow{4}{*}{f6} & link\_scope & "among" \\ \cline{2-3}
   & leaf\_value( ) & "the item that has the \{self.order\} largest \{self.h\} among all items" \\ \cline{2-3}
   & value( ) & "the item that has \{self.order\} largest \{self.h\}" \\ \cline{2-3}
   & question( ) & "which item has the \{self.order\} largest \{self.h\}" \\ \hline

\multirow{4}{*}{f7} & link\_scope & "of" \\ \cline{2-3}
   & leaf\_value( ) & "the \{self.h\} of \{self.row\}" \\ \cline{2-3}
   & value( ) & "the \{self.h\}" \\ \cline{2-3}
   & question( ) & "what is the \{self.h\}" \\ \hline

 \multirow{2}{*}{f8}  & \par value \par(val, val2=self.val2) & "\{val\} plus \{val2\}" \\ \cline{2-3}
   & \par question \par(val, val2=self.val2) & "what is \{val\} plus \{val2\}" \\ \hline

 \multirow{2}{*}{f9}  & leaf\_value( ) & "its \{self.h\} plus \{self.val\}" \\ \cline{2-3}
   & value(val=self.val) & "its \{self.h\} plus \{val\}" \\ \hline

\end{tabular}

\caption{\label{tab:class} Classes for each primitive. Attributes are without parentheses, functions are with parentheses, and arguments with an equal sign mean that the value to the right of the equal sign is used by default. Values starting with "self." are randomly generated parameters within the class.}
\end{table*}

\begin{table*}[h]

\centering

\begin{tabular}{p{0.1\textwidth}|p{0.8\textwidth}}

\hline
\textbf{Type} & \textbf{Template} \\ \hline
0-AAA & q = "Among \{f1.scope( )\} \{f1.link\_scope\} \{f0.leaf\_scope( )\}, \{f3.question( )\} \{f3.link\_scope\} \{f2.scope( )\} among them?" \\ \hline
0-AAB & \par v = "\{f2.value( )\} \{f2.link\_scope\} \{f1.scope( )\} \{f1.link\_scope\} \{f0.leaf\_scope( )\}" \par q = "Among all items, \{f3.question(val=v)\}?" \\ \hline
0-ABA & \par v = "\{f1.value( )\} \{f1.link\_scope\} \{f0.leaf\_scope( )\}" \par q = "\{f3.link\_scope\} \{f2.scope(val=v)\}, \{f3.question( )\}?" \\ \hline
0-ABB & \par v1 = "\{f1.value( )\} \{f1.link\_scope\} \{f0.leaf\_scope( )\}" \par v2 = "\{f2.value(val=v1)\}" \par q = "Among all items, \{f3.question(val=v2)\}?" \\ \hline
0-BAA & q = "Among \{f1.scope(val=f0.leaf\_value( ))\}, \{f3.question( )\} \{f3.link\_scope\} \{f2.scope( )\} among them?" \\ \hline
0-BAB & \par v = "\{f2.value( )\} \{f2.link\_scope\} \{f1.scope(val=f0.leaf\_value( ))\}" \par q = "Among all items, \{f3.question(val=v)\}?" \\ \hline
0-BBA & \par v = "\{f1.value(val=f0.leaf\_value( ))\}" \par q = "\{f3.link\_scope\} \{f2.scope(val=v)\}, \{f3.question( )\}?" \\ \hline
0-BBB & q = "Among all items, \{f3.question(val=f2.value(val=f1.value(val=f0.leaf\_value( ))))\}?" \\ \hline
1-ABA & \par q = "Among \{f0.leaf\_scope( )\}, \{f3.question( )\} \{f3.link\_scope\} \par\{f2.scope(val=f1.leaf\_value( ))\}?" \\ \hline
1-BBB & \par q = "Among all items, \{f3.question(val=f2.value(val=f0.leaf\_value( ), \par val2=f1.leaf\_value( )))\}?" \\ \hline
2-AAB & \par q = "Among \{f1.scope( )\} \{f1.link\_scope\} \{f0.leaf\_scope( )\}, \par\{f3.question(val=f2.leaf\_value( ))\}?" \\ \hline
2-ABA & \par v = "\{f1.value( )\} \{f1.link\_scope\} \{f0.leaf\_scope( )\}" \par q = "Among \{f2.leaf\_scope( )\}, \{f3.question(val=v)\}?" \\ \hline
2-ABB & \par v = "\{f1.value( )\} \{f1.link\_scope\} \{f0.leaf\_scope( )\}" \par q = "\{f3.question(val=v, val2=f2.leaf\_value( ))\}?" \\ \hline
2-BBA & \par v = "\{f1.value(val=f0.leaf\_value( ))\}" \par q = "Among \{f2.leaf\_scope()\}, \{f3.question(val=v)\}?" \\ \hline
3-AAB & \par v = "\{f2.value( )\} \{f2.link\_scope\} group A" \par q = "Assuming that group A contains \{f0.leaf\_scope( )\}, \{f3.question(val=v)\} \{f3.link\_scope\} \{f1.scope( )\} among group A?" \\ \hline
3-ABB & \par v1 = "\{f1.value( )\} \{f1.link\_scope\} them" \par v2 = "\{f2.value( )\} \{f2.link\_scope\} them" \par q = "Among \{f0.leaf\_scope( )\}, \{f3.question(val=v1, val2=v2)\}?" \\ \hline
4-AAB & \par v = "\{f2.value( )\} \{f2.link\_scope\} them" \par q = "Among \{f1.scope( )\} \{f1.link\_scope\} \{f0.leaf\_scope( )\}, \{f3.question(val=v)\}?" \\ \hline
5-ABA & \par v = "\{f1.value( )\} \{f1.link\_scope\} them" \par q = "Among \{f0.leaf\_scope( )\}, \{f3.question( )\} \{f3.link\_scope\} \{f2.scope(val=v)\} among them?" \\ \hline

\end{tabular}

\caption{\label{tab:template} Question generation templates for different types of schemes. The first number of the type is the number of the base graph (0 to 5 in left-to-right order in Figure~\ref{fig:base_graph}). The nodes in the graph are labeled 0 to 3 in order from bottom to top and from left to right at the same height. The letter in the type represents the type of the node, with A referring to nodes with the scope function and B referring to nodes with the value function.}
\end{table*}

We first design classes for each primitive, containing attributes and callable functions. See Table~\ref{tab:class} for classes for each primitive. For different types of schemes, we design templates to generate problems using the node's primitive class. See Table~\ref{tab:template} for question generation templates for different types of schemes.

\subsection{Answer Checking}
\label{app:answer}

The execution of variable substitution requires a storage containing the mapping pairs $(x, y)$. For each expression $L := R$ in turn, each $x$ in the storage in $R$ is replaced with the corresponding $y$, and then $(L,R)$ is added to the storage. The $y$ corresponding to the "result" in the storage is used as the final single expression.

When there is a primitive $f$ with swappable arguments (currently only f8), both variable substitutions $L:=f(A, B)$ and $L:=f(B, A)$ are taken into account, and branches are generated. $2^k$ final expressions are checked if there are $k$ primitives with swappable arguments in the expressions.

\section{Discussions}

\subsection{Grammatical Divergence}
\label{app:gram}
There are potential issues of grammatical divergence in the question generation template. For example, in the first clause of the question shown on the left-side of Figure~\ref{fig:framework}, we use "it" to refer to each of the individuals in "items" and use the phrase "satisfy [a sentence]", which diverges from common grammar. It is more common to use "whose" to lead the following clause "attr\_958 is...". We re-experiment on DeepSeek after changing all the parts of the template associated with this divergence to use "whose", resulting in a small change in performance ($P_c -0.31$ and $P_{dc} + 0.31$). We hypothesize that grammatical divergences that do not affect comprehension have a very limited impact on the results, especially given that the samples in the demonstrations use the same grammar as the test sample. Therefore, we still maintain the parts with potential grammatical divergences.

\subsection{Framework Generalizability}
\label{app:gen}
The introduction of decomposition and the decoupling of composition and decomposition in the DEDC framework can be easily generalized to other N2F tasks. After identifying the considered primitives and the names that hide the meaning, the evaluation can be performed as described in the framework. In contrast, the generalization of graph-based sample construction needs consideration of the degree of fitness to the task and will necessarily require new question template designs. Nevertheless, due to the independence of the sample and task construction, the DEDC framework can still be generalized directly to tasks that are already supported by samples. For extra settings, counter-intuitive symbolic names can be easily generalized to any task, while the generalization of compositional gaps requires consideration of which combinatorial pattern on the task is of concern. Furthermore, other additional settings are still to be explored.

\subsection{Human Capabilities}
\label{app:human}

Persons familiar with programming are expected to accomplish the task described in this work simply by following the methodology of (1) completing the decomposition based on the indication of the position of the parameters, (2) utilizing familiarity with the functions to complete the composition and cope with the compositional gaps, and (3) for counter-intuitive symbolic names, mapping the names to the normal names first, and reflecting the mappings after the processing is complete. However, we are currently unable to conduct large-scale human experiments to determine human capabilities on this task. There is also a lack of effective means to explore the similarities and differences between LLMs and humans in processing this task. Human-related investigations from a cognitive perspective are still to be explored.

\end{document}